\def\year{2021}\relax
\documentclass[letterpaper]{article} % DO NOT CHANGE THIS
\usepackage{aaai21}  % DO NOT CHANGE THIS
\usepackage{times}  % DO NOT CHANGE THIS
\usepackage{helvet} % DO NOT CHANGE THIS
\usepackage{courier}  % DO NOT CHANGE THIS
\usepackage[hyphens]{url}  % DO NOT CHANGE THIS
\usepackage{graphicx} % DO NOT CHANGE THIS
\usepackage{natbib}  % DO NOT CHANGE THIS AND DO NOT ADD ANY OPTIONS TO IT
\usepackage{caption} % DO NOT CHANGE THIS AND DO NOT ADD ANY OPTIONS TO IT
\usepackage{adjustbox}
\usepackage{amsmath,amssymb}
\usepackage{booktabs}
\usepackage{epsfig}
\usepackage{multirow}
\usepackage{graphics}
\usepackage{subcaption}
\title{Towards a Better Understanding of VR Sickness: \\ Physical Symptom Prediction for VR Contents}
\author{
	Hak Gu Kim\textsuperscript{\rm 1,2}\thanks{Work done as a part of the research project in KAIST},
	Sangmin Lee\textsuperscript{\rm 1},
	Seongyeop Kim\textsuperscript{\rm 1},
	Heoun-taek Lim\textsuperscript{\rm 1},
	Yong Man Ro\textsuperscript{\rm 1}\thanks{Corresponding author (ymro@kaist.ac.kr)}
	\\
}
\affiliations{
	\textsuperscript{\rm 1}Image and Video Systems Lab., KAIST, Korea\\
	\textsuperscript{\rm 2}School of Computer and Communication Sciences, EPFL, Switzerland
	\\
	hakgu.kim@epfl.ch, \{sangmin.lee, seongyeop, ingheoun, ymro\}@kaist.ac.kr
}

\begin{document}

\maketitle

\begin{abstract}
We address the black-box issue of VR sickness assessment (VRSA) by evaluating the level of physical symptoms of VR sickness. For the VR contents inducing the similar VR sickness level, the physical symptoms can vary depending on the characteristics of the contents. Most of existing VRSA methods focused on assessing the overall VR sickness score. To make better understanding of VR sickness, it is required to predict and provide the level of major symptoms of VR sickness rather than overall degree of VR sickness. In this paper, we predict the degrees of main physical symptoms affecting the overall degree of VR sickness, which are disorientation, nausea, and oculomotor. In addition, we introduce a new large-scale dataset for VRSA including 360 videos with various frame rates, physiological signals, and subjective scores. On VRSA benchmark and our newly collected dataset, our approach shows a potential to not only achieve the highest correlation with subjective scores, but also to better understand which symptoms are the main causes of VR sickness.
\end{abstract}

%===================================================================================
%								1. INTRODUCTION		
%===================================================================================

\section{Introduction}

Virtual reality (VR) perception brings an immersive viewing experience to viewers with the development of commercial devices. However, there have been increasing unwanted side effects on the safety and health of the VR viewing. VR sickness, which is one of cybersickness in VR environment \cite{2}, often occurs in many users exposed to VR content \cite{3}.

There are various physical symptoms that viewers perceive some levels of VR sickness. The possible physical symptoms of VR sickness have been extensively studied with respect to the viewing safety and health issues \cite{4,5}. In \cite{5}, through extensive subjective experiments and statistical analysis, three types of statistically significant symptoms were introduced: 1) disorientation, 2) nausea, and 3) oculomotor. To better understand and estimate VR sickness, it is required to investigate the levels of physical symptoms of VR sickness. But it is a non-trivial task due to a complex combination of various symptoms.

%---------------------------------------------Fig.1
%------------------------------------ Figure 1
\begin{figure}[t!]
\begin{center}
\includegraphics[width=0.9\linewidth] {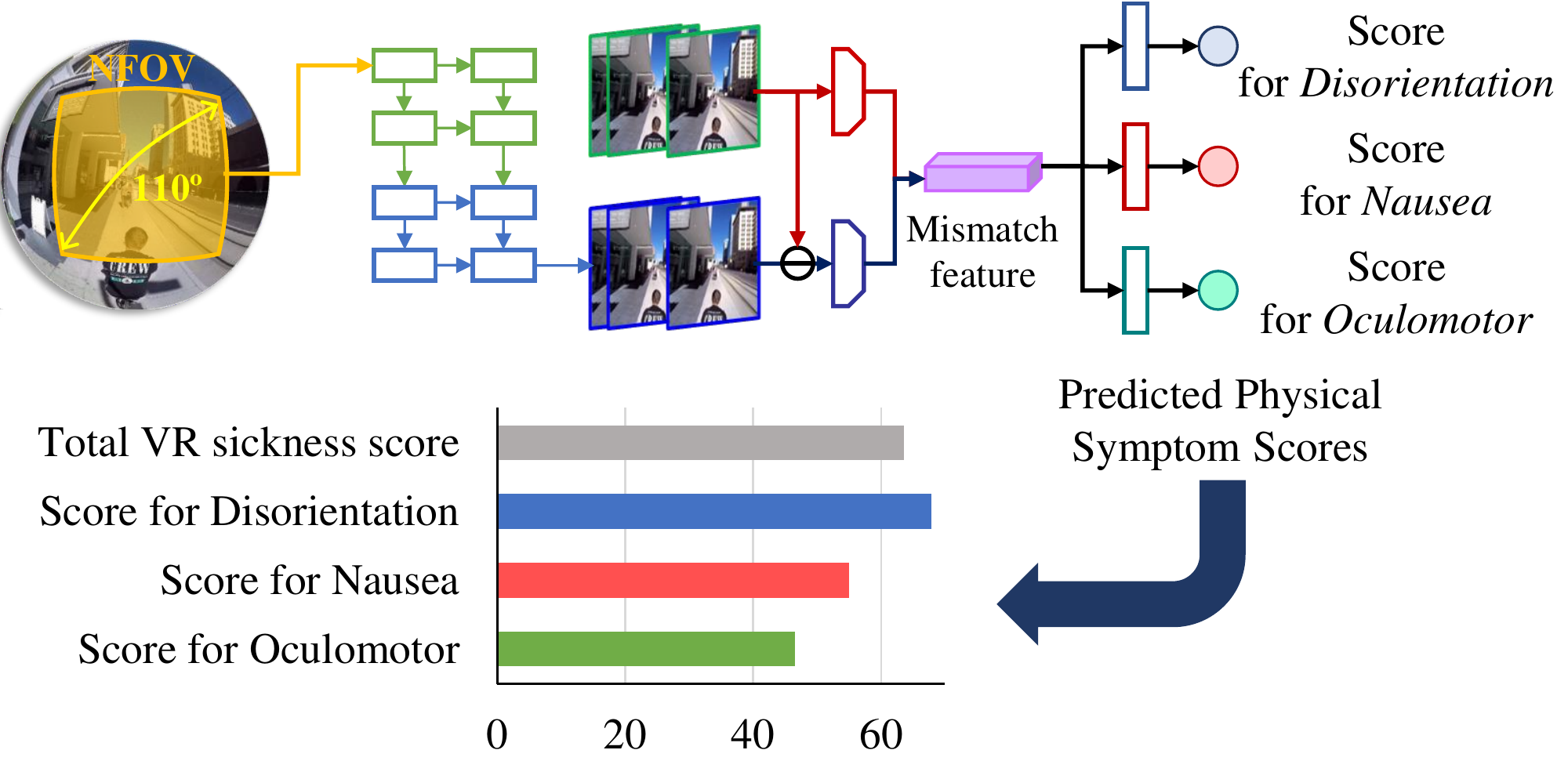}
\end{center}
   \caption{The intuition of physical symptom prediction for better understanding of VR sickness. In general, VR contents lead to different levels of physical symptoms according to their spatio-temporal characteristics.}
\label{fig:1}
\end{figure}

%--------------------------------------------------

To investigate and predict physical symptom responses of VR sickness, many existing works were proposed that measured various physiological signals from subjective assessment experiments or collected subjective questionnaires for physical symptoms of VR sickness through subjective experiments. However, it is time consuming and labor-intensive to conduct extensive subjective assessment experiments to obtain physiological signals and subjective questionnaires. Most recently, there were a few approaches for VR sickness assessment (VRSA) by analyzing the spatio-temporal characteristics of contents \cite{18,19} or both of contents and physiological signals \cite{34,21}. However, they just predicted level of total VR sickness score or simple mean opinion score (MOS), instead of physical symptoms responses.

To address that, we firstly propose a novel physical symptom prediction to understand the physical symptoms caused by VR sickness and make better understanding of VR sickness. Potentially, a reliable physical symptom prediction can be applicable to a wide range of automation services in VR content generation and platform services. For example, in Oculus, users select and experience 360-degree contents they want to see with reference to its VR sickness level. In VR content selection, one important problem is that VR contents with similar sickness level may have different causes of VR sickness. Some users are vulnerable to disorientation caused by rapid rotation, whereas someone else may be tolerant of nausea caused by shaking, and vice versa. In this paper, as the first attempt to automatically investigate physical symptoms based on VR content analysis, this work aims at predicting the degrees of physical symptoms and investigating them for better understanding of VR sickness. For this purpose, we build a novel physical symptom prediction framework inspired by brain mechanism (see Fig.~\ref{fig:1}). By automatically evaluating each symptom score for disorientation, nausea, and oculomotor as well as a total VR sickness score, we can prevent users from experiencing severe VR sickness. This is because users can avoid watching the VR content causing them a vulnerable physical symptom. Our main contributions are as follows.

First, we propose to predict major physical symptom scores for understanding main causes of VR sickness, which are dependent on the spatio-temporal characteristics of 360-degree videos. Even if users perceive the similar level of overall VR sickness for different 360-degree videos, the perceived physical symptoms (i.e., the causes of VR sickness) could vary due to the content characteristics. In this paper, we estimate not only the total VR sickness score but also the symptom scores for disorientation, nausea, and oculomotor.

Second, we design a novel objective physical symptom prediction method considering neural mismatch mechanism in order to reliably predict each level of  three major physical symptoms. The neural mismatch mechanism is a widely accepted theory of VR sickness \cite{6,7,8}. The converging sensory inputs from the visual sensor (eyes) are compared with the expected sensory signals by neural store (brain), which are calibrated by past experience. Then, the discrepancy between the current sensory inputs and the expected sensory patterns leads to produce mismatches (i.e., neural mismatch signal). Finally, physical symptoms of VR sickness can be activated when the mismatch signal is more excessive than the tolerance of human perception. Inspired by this mechanism, the proposed framework consists of neural store network, comparison network, and physical symptom score prediction network. In the neural store network, the next frame (i.e., expected visual signal) is predicted from input frames based on the trained parameters in training. Similar to our experience in daily life, by training the neural store network with the videos that might not cause severe VR sickness, it can learn the spatio-temporal characteristics of the normal visual signals. The comparison network is to encode the mismatches between input and the expected frames. By encoding the discrepancy between them, the mismatch feature can be encoded in the comparison network. Finally, in the physical symptom score prediction network, three main symptom scores for disorientation, nausea and oculomotor are evaluated from the encoded mismatch feature.

Third, for the evaluation, we collect a new large-scale dataset for VR physical symptom prediction (VRPS) that includes eighty 360-degree videos with four different frame rates, subjects’ physiological signals (heart rate and galvanic skin conductance), and the corresponding subjective physical symptom scores. To collect a large-scale VRPS dataset, we conduct extensive subjective assessment experiments for encouraging physical symptom prediction research fields.

Experimental results show that this study can provide more meaningful perception information by estimating major physical symptoms as well as total VR sickness. In particular, we demonstrate our model lets users know what physical symptoms can be induced for a given content.

We summarize the contributions of this work as follows.
\begin{itemize}
  \item To the best of our knowledge, we propose a first physical symptom score prediction approach for VRSA. We introduce and predict major physical symptoms of VR sickness for disorientation, nausea, and oculomotor.
  \item We propose a novel objective assessment based on a neural mismatch mechanism. Our model consists of (i) neural store network for expecting visual signals from past experience, (ii) comparison network for encoding the discrepancy between input and the expected visual signals, (iii) physical symptom score prediction network for estimating the levels of three physical symptoms.
  \item For the performance evaluation, we conduct extensive subjective assessment experiments and introduce a large-scale physical symptom prediction benchmark database. We make it publicly available on the Web\footnote{http://ivylabdb.kaist.ac.kr}.
\end{itemize}

%===================================================================================
%								2. PROPOSED METHOD
%===================================================================================

\section{Related Work}
\textbf{VRSA using physiological measurement and subjective questionnaire} Many existing works focused on the subjective assessment studies using physiological measurements \cite{9,10,11} and subjective questionnaires \cite{5,12,13,14,15}. The authors of \cite{9} investigated to the relation between the changes in a variety of physiological signals and simulator sickness questionnaires (SSQ) scores obtained by subjects. They measured electroencephalography (EEG), electrogastrogram (EGG), galvanic skin response (GSR), $etc$. The experimental results showed that the changes in the activity of the central and autonomic nervous systems had a positive correlation with VR sickness. In \cite{15}, subjective studies were conducted to measure the quality of experience (QoE) and VR sickness of 360-degree videos by assessing MOS and SSQ score, respectively. However, the approaches using the physiological measurements and subjective questionnaires were very cumbersome and labor-intensive. It cannot prevent viewers from watching VR contents causing severe VR sickness.

%---------------------------------------------Fig.2
%------------------------------------ Figure 2
\begin{figure*}[t!]
\begin{center}
\includegraphics[width=1.0\linewidth] {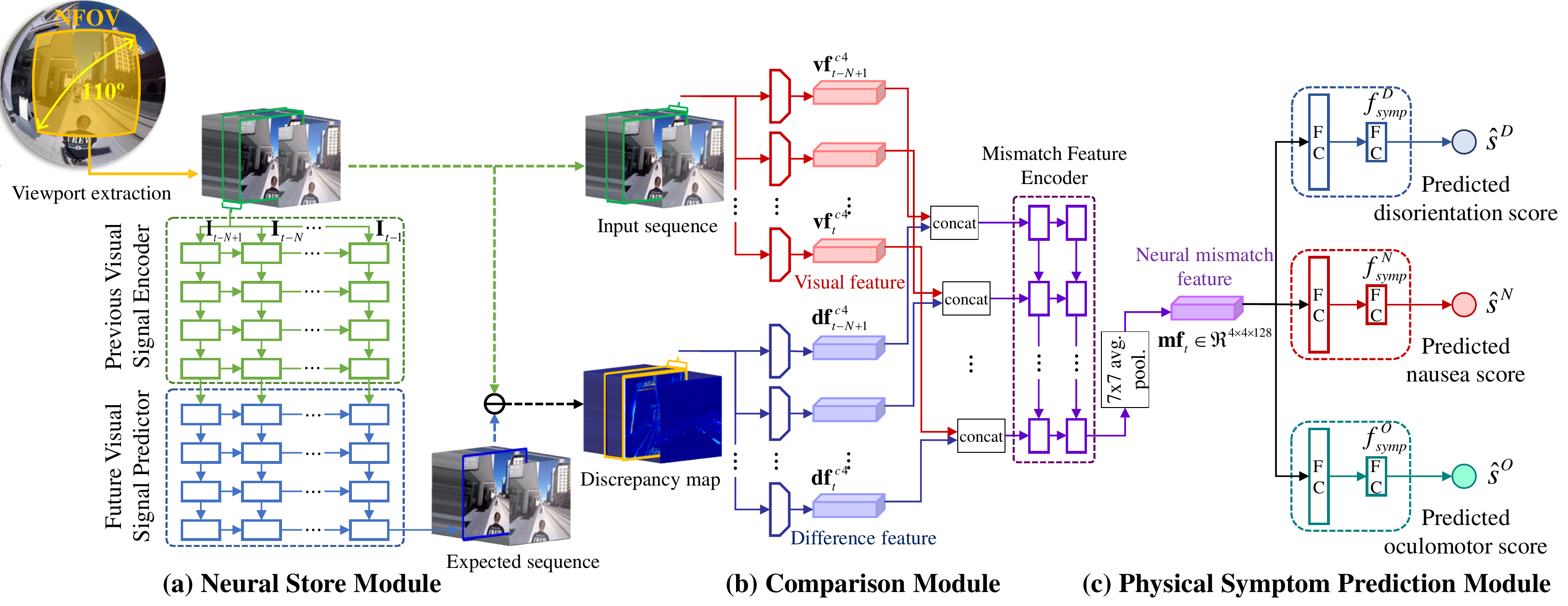}
\end{center}
   \caption{The illustration of our physical symptom prediction considering neural mismatch mechanism.}
\label{fig:2}

\end{figure*}

%--------------------------------------------------

\textbf{VRSA using content analysis} There were a few content analysis-based VRSA methods using machine learning techniques \cite{16,17,18,19,20,21}. In \cite{16}, a VR sickness predictor based on perceptual motion feature and statistical content feature was presented using support vector regression (SVR). In \cite{17}, a decision tree-based sickness predictor for 360-degree stereoscopic videos was proposed using disparity and optical flow to estimate the nauseogenicity of VR content. In \cite{18,19}, the deep learning-based VRSA methods were proposed considering exceptional motion of VR content. In \cite{20}, a new motion sickness prediction model was proposed using 3D convolutional neural networks (CNNs) for stereoscopic videos.  However, they could not provide which physical symptoms have a significant effect on users’ VR sickness due to ‘black-box’ regression between the content feature and the final subjective sickness score. Therefore, it is essential to estimate the degrees of physical symptoms as well as the overall VR sickness level.

\textbf{VRSA using content information and physiological signals} Most recently, objective assessment methods considering both content information and physiological signals have been proposed \cite{34,21,36}. In \cite{34,36}, the authors proposed a deep learning-based individual VR sickness assessment method considering content stimulus and physiological responses of each subject. Experimental results showed that this approach was effective to predict the level of individual VR sickness based on individual physiological signals. A deep cybersickness predictor was proposed considering brain signal analysis \cite{21}. Unlike them, we introduce a content analysis-based physical symptom prediction for more practical VR applications.

%===================================================================================
%								3. PROPOSED METHOD
%===================================================================================

\section{Physical Symptom Prediction for Understanding VR Sickness}

Figure~\ref{fig:2} shows the proposed physical symptom prediction inspired by human VR perception, a neural mismatch mechanism. In this work, we focus on the visual sensory signals based on content analysis. Let $\textup{\textbf{V}}$ and $\hat{\textup{\textbf{V}}}$ denote the original video and the expected video, respectively. $\textup{\textbf{mf}}$ denotes the neural mismatch feature, which learns the discrepancy between $\textup{\textbf{V}}$ and $\hat{\textup{\textbf{V}}}$. $\hat{s}^{D}, \hat{s}^{N},$ and $\hat{s}^{O}$ are the predicted symptom scores for disorientation, nausea, and oculomotor, respectively.

Overall, training our physical symptom prediction model consists of two steps: 1) training of the neural store network and 2) training of both comparison network and physical symptom score prediction network. In the training, the neural store network learns the visual signal expectation from the input signal. To learn our experience in daily life, the neural store network is trained with normal video dataset that has non-exceptional motion patterns and high frame rate (i.e., these characteristics could not lead to VR sickness). In the testing, each frame is expected by the trained neural store network. Since the neural store network is trained with normal videos that could not lead to VR sickness, it could well-predict the normal videos with non-exceptional motion patterns and high frame rate. On the other hand, the 360-degree videos with exceptional motion patterns (i.e., acceleration and rapid rotation) or low frame rate cannot be predicted well. With the context and difference information between input and the expected frames, the symptom scores for disorientation, nausea, and oculomotor are predicted by mapping the mismatch feature onto each physical symptom score. In addition, an overall score for VR sickness is calculated from the predicted physical symptom scores. In this paper, the SSQ scores for disorientation, nausea, and oculomotor obtained by subjects are used as a ground-truth physical symptom scores.

\subsection{Neural Store Network}
The proposed neural store network consists of the previous visual signal encoder, the future visual signal predictor for expectation of next frame and the spatio-temporal discriminator via adversarial learning \cite{1}. Since people generally cannot experience the situation causing severe VR sickness such as exceptional motion, shaking and low frame rate, the neural store network is also trained with normal video sequence. In this paper, normal videos mean the contents do not cause severe VR sickness and their total SSQ scores are under about 30. In general, they involve non-exceptional motion (static or slow driving) or high frame rate (over 30 fps). On the other hand, the videos leading to severe VR sickness (e.g., roller coaster and racing) have a total SSQ score of over 30. To teach the neural store network the general experience of people, we use the normal videos without acceleration, rapid rotation, shaking, $etc$. In training stage, the discriminator takes the original video or the expected video. Then, it determines whether a given video has a distribution of the normal video or not.

In the proposed method, pleasantly looking normal field of view (NFOV) segments from infinite FOV of 360-degree videos are used as input frames \cite{19}. The NFOV can be obtained by equirectangular projection with the viewpoint as a center. The size of the NFOV region is set to span 110-degree diagonal FOV \cite{22}, same as that of the high-end HMD. Let $\textup{\textbf{I}}_{t}$ and $\hat{\textup{\textbf{I}}}_{t}$ denote $t$-th input frame and $t$-th predicted frame, respectively. $\textup{\textbf{R}}_{t}$ and $\textup{\textbf{F}}_{t}$ denote a set of original NFOV video frames (i.e., $\textup{\textbf{R}}_{t} = [\textup{\textbf{I}}_{t-N}, \cdots, \textup{\textbf{I}}_{t}]$) and a fake NFOV video sequence (i.e., $\textup{\textbf{F}}_{t} = [\hat{\textup{\textbf{I}}}_{t-N}, \cdots, \hat{\textup{\textbf{I}}}_{t}]$), respectively ($N=10$).
The proposed previous visual signal encoder and future visual signal predictor consist of convolutional LSTM (ConvLSTM) and deconvolutional LSTM (DeconvLSTM) \cite{23}, respectively. The encoder and predictor are composed of 4 layers of ConvLSTM and 4 layers of DeconvLSTM, respectively. All layers have $3\times3$ filter with stride (2,2). The $t$-th predicted frame can be defined as
%----------------------------------------------------------------------------------------------------------------------Eq. 1
\begin{equation}
	\label{eq:1}
		\hat{\textup{\textbf{I}}}_{t} = {P}_{\theta}(\textup{\textbf{I}}_{t-N}, \cdots, \textup{\textbf{I}}_{t-1}),
\end{equation}
%---------------------------------------------------------------------------------------------------------------------------
where ${P}_{\theta}$ means the neural store network with parameters $\theta$.

Through adversarial learning, the predictor predicts the next normal video frame well in order to deceive the discriminator. To that end, we design the loss function of the encoder and predictor using prediction loss and realism loss \cite{35}. By minimizing the prediction loss between the original frame $\textup{\textbf{I}}_{t}$ and the expected frame $\hat{\textup{\textbf{I}}}_{t}$, the prediction quality can be enhanced. The realism loss helps the predicted frame to be realistic enough to fool the discriminator. Finally, total loss of the proposed neural store network, $L_P$, can be defined as a combination of the prediction loss and the realism loss.
%----------------------------------------------------------------------------------------------------------------------Eq. 2
\begin{equation}
	\label{eq:2}
	\begin{split}
		L_P(\theta, \phi) = &{\|{P}_{\theta}(\textup{\textbf{I}}_{t-N}, \cdots, \textup{\textbf{I}}_{t-1}) - \textup{\textbf{I}}_{t}) \|}^2_2 \\
							& - {\lambda}_{a}\log({D}_{\phi}(\textup{\textbf{F}}_{t})),
	\end{split}
\end{equation}
%---------------------------------------------------------------------------------------------------------------------------
where ${D}_{\phi}$ indicates the discriminator with parameters ${\phi}$. ${\lambda}_{a}$ is a weight parameter to control the balance between the first term for prediction and the second term for realism. $\textup{\textbf{F}}_{t}$ is a fake sequence including the predicted frame.

The discriminator is to determine whether the input is realistic or not by considering its spatio-temporal characteristics. The spatio-temporal discriminator is based on the 3-D CNN, which contains 5 layers and 64-d fully-connected (FC) layer. All 3D kernels are $3\times3\times3$ with stride (1,2,2). Our discriminator loss, ${L_D}$, can be written as 
%----------------------------------------------------------------------------------------------------------------------Eq. 3
\begin{equation}
	\label{eq:3}
		L_D(\phi) = \log(1-{D}_{\phi}(\textup{\textbf{F}}_{t})) + \log({D}_{\phi}(\textup{\textbf{R}}_{t})).
\end{equation}
%---------------------------------------------------------------------------------------------------------------------------

In Eq. (3), the ${D}_{\phi}(\textup{\textbf{F}}_{t})$ in the first term is the probability that the discriminator determines the fake sequence as original. The second term, ${D}_{\phi}(\textup{\textbf{R}}_{t})$, is the probability that the discriminator determines the real sequence as original.

For training the neural store network including the discriminator with adversarial learning \cite{24}, we devise a new adversarial objective function, which can be written as
%----------------------------------------------------------------------------------------------------------------------Eq. 4
\begin{equation}
	\label{eq:4}
		\min_{P}\max_{D} V({P}_{\theta}, {D}_{\phi}):= L_P(\theta, \phi) + {\lambda}_{D}L_D(\phi),
\end{equation}
%---------------------------------------------------------------------------------------------------------------------------
where ${\lambda}_{D}$ is a weight parameter for the discriminator.

By the adversarial learning between ${P}_{\theta}$ and ${D}_{\phi}$, the prediction performance of the neural store network for normal video sequence can be enhanced.

\subsection{Comparison Network}
Our comparison network is designed to encode the mismatch information. After obtaining the expected videos by the trained neural store network, the difference information between original frame and the predicted frame is encoded as well as visual information of input video sequence. The discrepancy between the original frame (visual sensory input) and the expected frame, $\textup{\textbf{d}}_{t}$, can be defined as
%----------------------------------------------------------------------------------------------------------------------Eq. 5
\begin{equation}
	\label{eq:5}
		\textbf{d}_{t} = |\textup{\textbf{I}}_{t} - {P}_{\theta}(\textup{\textbf{I}}_{t-N}, \cdots, \textup{\textbf{I}}_{t-1})|=|\textup{\textbf{I}}_{t} - \hat{\textup{\textbf{I}}}_{t}|.
\end{equation}
%---------------------------------------------------------------------------------------------------------------------------

The discrepancy map, $\textup{\textbf{d}}_{t}$, indicates the gap between input frame (i.e., visual signal sensed by our eye) and the expected frame by neural store network (i.e., visual signal expected from our experience in neural store). The visual information is also important factors affecting VR sickness. In the proposed method, we take into account both visual information and the difference between the original and the predicted videos in order to correctly predict the levels of physical symptoms. In this study, the feature map of 4-th convolutional layers of VGG-19 \cite{25} is used as the visual feature $\textbf{vf}^{c4}\in \mathbb{R}^{56\times 56\times 128}$ and the difference feature $\textbf{df}^{c4}\in \mathbb{R}^{56\times 56\times 128}$. $\textbf{mf}\in \mathbb{R}^{4\times 4\times 128}$ denotes the neural mismatch feature. After obtaining $\textbf{vf}^{c4}$ and $\textbf{df}^{c4}$, they are concatenated as $[\textbf{vf}^{c4}; \textbf{df}^{c4}]$. Then, $[\textbf{vf}^{c4}; \textbf{df}^{c4}]$ is encoded to produce the mismatch feature $\textbf{mf}$ using 3 layers of ConvLSTM with $3\times3$ filter and stride (2,2).

\subsection{Physical Symptom Score Prediction Network}
In the proposed physical symptom score prediction network, the degree of three major symptoms of VR sickness are predicted from the latent neural mismatch feature. Figure~\ref{fig:2}(c) shows the architecture of the proposed physical symptom score predictor. It consists of three fully connected layers, which are 64-dimensional, 16-dimensional, and 1-dimensional layers. The physical symptom score prediction network plays a role in non-linearly mapping the high-dimensional neural mismatch feature space onto the low-dimensional physical symptom score space. To that end, in training stage, the sickness scores for disorientation, nausea, and oculomotor are predicted from the latent neural mismatch feature $\textbf{mf}$ by minimizing the objective function between the predicted symptom scores and the ground-truth subjective scores. In this paper, mean SSQ score values for disorientation, nausea, and oculomotor obtained from subjects are used as the ground-truth physical symptom scores. The objective function for the physical symptom score prediction, $L_{symp}$, can be written as
%----------------------------------------------------------------------------------------------------------------------Eq. 6
\begin{equation}
	\label{eq:6}
	\begin{split}
		& L_{symp}=\frac{1}{K}\sum_{j=1}^{K} \left \{  \right. \left \| f_{symp}^{D}\left ( \textbf{mf} \right )-s_{j}^{D}\right \|^2_2 +\\ 
				 & \left \| f_{symp}^{N}\left ( \textbf{mf} \right )-s_{j}^{N}\right \|^2_2 + \left \| f_{symp}^{O}\left ( \textbf{mf} \right )-s_{j}^{O}\right \|^2_2 \left.  \right \},
	\end{split}
\end{equation}
%---------------------------------------------------------------------------------------------------------------------------
where $f_{symp}(\cdot)$ represents the non-linear regression by fully-connected layers for each symptom. $f_{symp}(\textbf{mf})$ indicates the predicted each symptom score. $s_{j}^{D}, s_{j}^{N}, and$ $s_{j}^{O}$ indicate the ground-truth subjective scores of $j$-th VR content for disorientation, nausea, and oculomotor, respectively. $K$ is the number of batches.

In testing stage, the expected video frames are obtained by the trained neural store network. Then, through the comparison network and physical symptom score prediction network, the physical symptom scores are obtained from the original and the expected sequences. In the testing, an overall degree of VR sickness (i.e., total VR sickness score, $\hat{s}^{VR}$) is estimated by weighted averaging the predicted physical symptom scores for disorientation, nausea, and oculomotor \cite{5}.

\begin{equation}
	\label{eq:7}
		\hat{s}^{VR}=3.74\times \left ( \frac{1}{13.92} \hat{s}^D + \frac{1}{9.54} \hat{s}^N + \frac{1}{7.58} \hat{s}^O \right ).
\end{equation}

%===================================================================================
%								4. BENCHMARK DATABASE
%===================================================================================

\section{Dataset for Physical Symptom Prediction}
To verify the effectiveness of our method, we built a new 360-degree video database and conducted extensive subjective assessment experiments to obtain the corresponding subjective symptom scores and physiological signals.

\subsection{360-Degree Video Dataset}
We collected a 4K 360-degree video datasets for performance evaluation of physical symptom score prediction named VRPS DB-FR. A total of twenty 360-degree videos were collected from Vimeo (see TABLE S1 in our supplementary file). They contain various normal driving with slow speed and slowly moving drone. Most of them have high frame rate of 30 fps or 60 fps. Thus, they might not lead to severe VR sickness caused by exceptional motion or low frame rate factors. From the twenty 360-degree videos, a total of 80 videos were generated using the optical flow interpolation of Adobe Premiere with the target 4 different frame rates, which are 10, 15, 30, and 60 fps (80 stimuli = 20 videos $\times$ 4 different frame rates). To match the video length, the frames of each video with 10, 15, and 30 fps were repeated 6, 4, and 2 times, respectively. As a result, the total number of frames for each video is the same (i.e., 3,600 frames during 60 sec.). Due to the viewing safety of the participated subjects, each video was presented for 60 sec.

%---------------------------------------------Fig.3
%------------------------------------ Figure 3
\begin{figure}[t!]
\begin{center}
\includegraphics[width=0.80\linewidth] {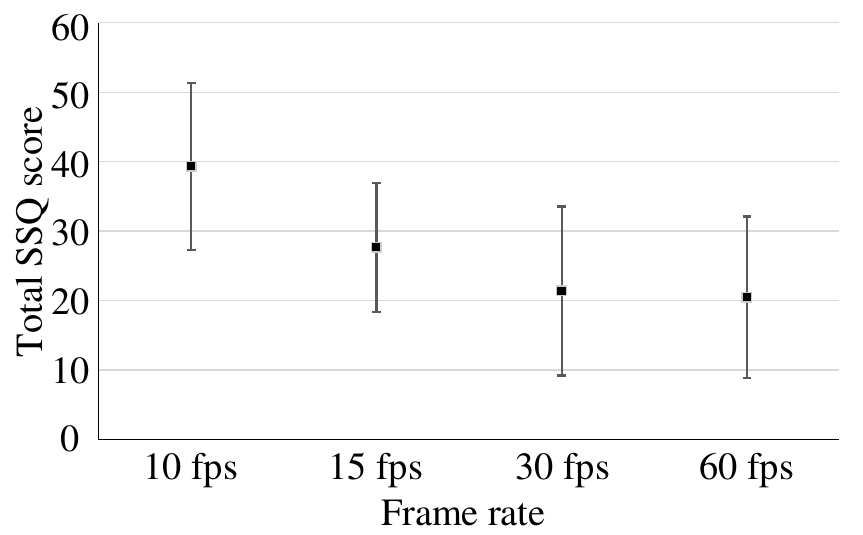}
\end{center}
   \caption{Subjective assessment results of VR sickness for the 360-degree video dataset with different frame rates.}
\label{fig:3}

\end{figure}

%--------------------------------------------------

\subsection{Subjective Assessment Experiment}
In subjective assessment experiments, Oculus Rift CV1 was used for displaying 360-degree videos, which was one of the high-end HMDs. Its resolution was $2160 \times 1200$ pixels ($1080 \times 1200$ pixels per eye). Its display frame rate was maximum 90 fps and it had 110-degree diagonal FOV.

A total of twenty subjects, aged 20 to 30, participated in our subjective experiment under the approval of KAIST Institutional Review Board (IRB). They had normal or corrected-to-normal vision. Before watching each stimulus, they were placed in the center to be started from zero position. In our experiments, their head motion was negligible during watching 360-degree contents because most of them were focusing their gaze in the camera movement direction \cite{26}. All experimental environments followed the guideline of recommendations of ITU-R BT.500-13 \cite{27} and BT.2021 \cite{28}.

Each video was randomly displayed for 60 sec. The resting time was given as 90 sec with mid gray image. During the resting time, subjects were asked to estimate the level of perceived VR sickness using 16-item SSQ score sheet \cite{5,29}. Each subject took about 50 min to complete one session for 20 stimuli. Total four sessions were conducted to complete 80 stimuli for each subject. Each session was performed in a different day. During the experiment, subjects were allowed to immediately stop and take a break if they felt difficult to continue the experiment due to excessive VR sickness.

In addition, we measured GSR and heart rate (HR) of subjects in our experiment. HR and GSR were measured using NeuLog heart rate/pulse sensor (NUL-208) and GSR sensor (NUL-207), respectively. The heart rate/pulse sensor consisted of an infrared LED transmitter and a matched infrared photo transistor receiver. The GSR sensor consisted of two probes and finger connectors. Sampling rate was 100 fps.

\subsection{Subjective Assessment Results}
Figure~\ref{fig:3} shows the distribution of VR sickness scores for VRPS DB-FR. The x-axis and y-axis indicate the frame rate and total SSQ score (i.e., VR sickness score), respectively. The black dot indicates the mean SSQ scores for stimuli at each frame rate. The SSQ scores of the 360-degree videos with high frame rates (30 fps and 60 fps) were low. Note that the total SSQ score ranging of 30 to 40 indicates noticeable VR sickness \cite{9}. On the other hand, the total SSQ scores of the VR contents with low frame rates (10 fps and 15 fps) were higher than those of high frame rate contents. The result is consistent with previous studies \cite{12,30}.

%===================================================================================
%								5. EXPERIMENTS AND RESULTS
%===================================================================================

%-------------------------------------------Table 1
% \usepackage{multirow}
\begin{table*}[t!]
\begin{adjustbox}{width=1\textwidth}
\centering
\begin{tabular}{c|c|c|c|c|c|c|c|c|c|c}
\hline
\hline
\multirow{2}{*}{DB}           & \multirow{2}{*}{Method}    & \multicolumn{3}{c|}{Disorientation} & \multicolumn{3}{c|}{Nausea} & \multicolumn{3}{c}{Oculomotor} \\ \cline{3-11} 
                              &                            & PLCC       & SROCC     & RMSE       & PLCC    & SROCC   & RMSE    & PLCC     & SROCC    & RMSE      \\ \hline
\multirow{3}{*}{KAIST IVY DB} & \cite{18} & 0.497      & 0.521     & 12.132     & 0.387   & 0.401   & 15.311  & 0.432     & 0.448      & 13.912    \\ \cline{2-11} 
							  & OF+PSP    & 0.661      & 0.621     & 11.337     & 0.556   & 0.531   & 13.726  & 0.538     & 0.522      & 13.745    \\ \cline{2-11}                              
                              & NS+PSP (ours)               & 0.701      & 0.658     & 10.271     & 0.667   & 0.631   & 12.866  & 0.678    & 0.656    & 11.174    \\ \cline{2-11} 
                              & NS+C+PSP (ours)             & \textbf{0.923}      & \textbf{0.921}     & \textbf{9.281}      & \textbf{0.875}   & \textbf{0.872}   & \textbf{9.792}   & \textbf{0.871}    & \textbf{0.853}    & \textbf{9.881}     \\ \hline
\multirow{3}{*}{VRPS DB-FR}   & \cite{18} & 0.352      & 0.354     & 15.226     & 0.334   & 0.330   & 13.492  & 0.343     & 0.320      & 13.489    \\ \cline{2-11} 
							  & OF+PSP    & 0.588      & 0.565     & 12.757     & 0.520   & 0.511   & 13.152  & 0.503     & 0.491      & 13.128    \\ \cline{2-11}                              
                              & NS+PSP (ours)               & 0.654      & 0.621     & 10.021     & 0.633   & 0.605   & 11.087  & 0.631    & 0.609    & 10.985    \\ \cline{2-11} 
                              & NS+C+PSP (ours)             & \textbf{0.842}      & \textbf{0.829}     & \textbf{6.821}      & \textbf{0.831}   & \textbf{0.809}   & \textbf{7.111}   & \textbf{0.821}    & \textbf{0.801}    & \textbf{7.405}     \\ \hline
                              \hline
\end{tabular}
\end{adjustbox}
\caption{Physical Symptom Level Prediction Performance on Two VRSA Databases}
\end{table*}
%--------------------------------------------------

\section{Experiments and Results}

\subsection{Experimental Setting and Network Training}

To verify the performance of the proposed physical symptom prediction model, experiments were conducted with two benchmark datasets, which are KAIST IVY 360-degree video dataset with different motion patterns \cite{1,19} and our dataset with different frame rates.

For training the neural store network, we used various video datasets such as KITTI benchmark datasets \cite{31} and other 360-degree video contents collected from Vimeo (see TABLE VII in \cite{19}). In the experiment, they were used for pre-training of our neural store network.

For performance evaluation of the proposed physical symptom prediction on KAIST IVY 360-degree video dataset with different motion patterns \cite{1}, the proposed comparison and physical symptom score prediction networks were end-to-end trained by another twenty one 360-degree videos captured by photo experts (see TABLE VIII in \cite{19}). About 2,700 frames of each clip were used for training. For testing, nine 360-degree videos were used. (see TABLE III in \cite{19} for more details). For performance evaluation on our VRPS DB-FR, 5-cross validation was conducted. 64 video clips were used for training the comparison network and the physical symptom score prediction network, 16 video clips and the corresponding subjective scores were used for testing. 3,600 frames of each clip were used for training and testing.

The neural store network was pre-trained by 60 epochs with ADAM optimizer \cite{32}. We used a batch size of 3. For ADAM optimizer, the learning rate was initialized at 0.00005. $\beta 1$ and $\beta 2$ were set to 0.9 and 0.999, respectively. Weight decay was set to $10e-8$ per each iteration. The comparison and symptom score prediction networks were trained with the same settings.

\subsection{Prediction Performances}

For performance evaluation of the proposed symptom prediction for better understanding of VR sickness, we used two benchmark datasets consisting of the 360-degree videos, the corresponding SSQ scores and physiological signals (HR and GSR) of VR sickness. One is a publicly available dataset, which is KAIST IVY 360-degree video database with different motion patterns for VRSA \cite{1}. The other is our VRPS DB-FR with different frame rates. To evaluate the performance, we employed commonly used three measures which are Pearson linear correlation coefficient (PLCC), Spearman rank order correlation coefficient (SROCC), and root mean square error (RMSE). 

TABLE 1 shows the performance of our physical symptom score prediction on two different datasets. For performance comparison, we compared our symptom prediction with existing deep autoencoder-based VRSA method \cite{18}. ‘OF+PSP’ indicates that the optical flow map \cite{37} is fed to our physical symptom score prediction network to verify the effectiveness of the proposed neural store network. ‘NS+PSP’ indicates the neural store network + physical symptom score prediction network. For ‘NS+PSP’, the discrepancy map is directly mapped onto the physical symptoms by physical symptom prediction network. ‘NS+C+PSP’ indicates the neural store + comparison + symptom score prediction networks. In TABLE 1, our model could provide a reliable physical symptom prediction results. The optical flow map seemed to be effective for predicting disorientation scores. However, compared to the discrepancy map in this study, it was not good at predicting physical symptom because VR contents with complex motion patterns or rapid acceleration can lead to inaccurate motion estimation. Then, the inaccurate optical flow information can interfere with physical symptom score prediction. In particular, our model, ‘NS+C+PSP’, outperformed for physical symptom score estimation.

\begin{figure}[t!]
\begin{center}
\includegraphics[width=0.9\linewidth] {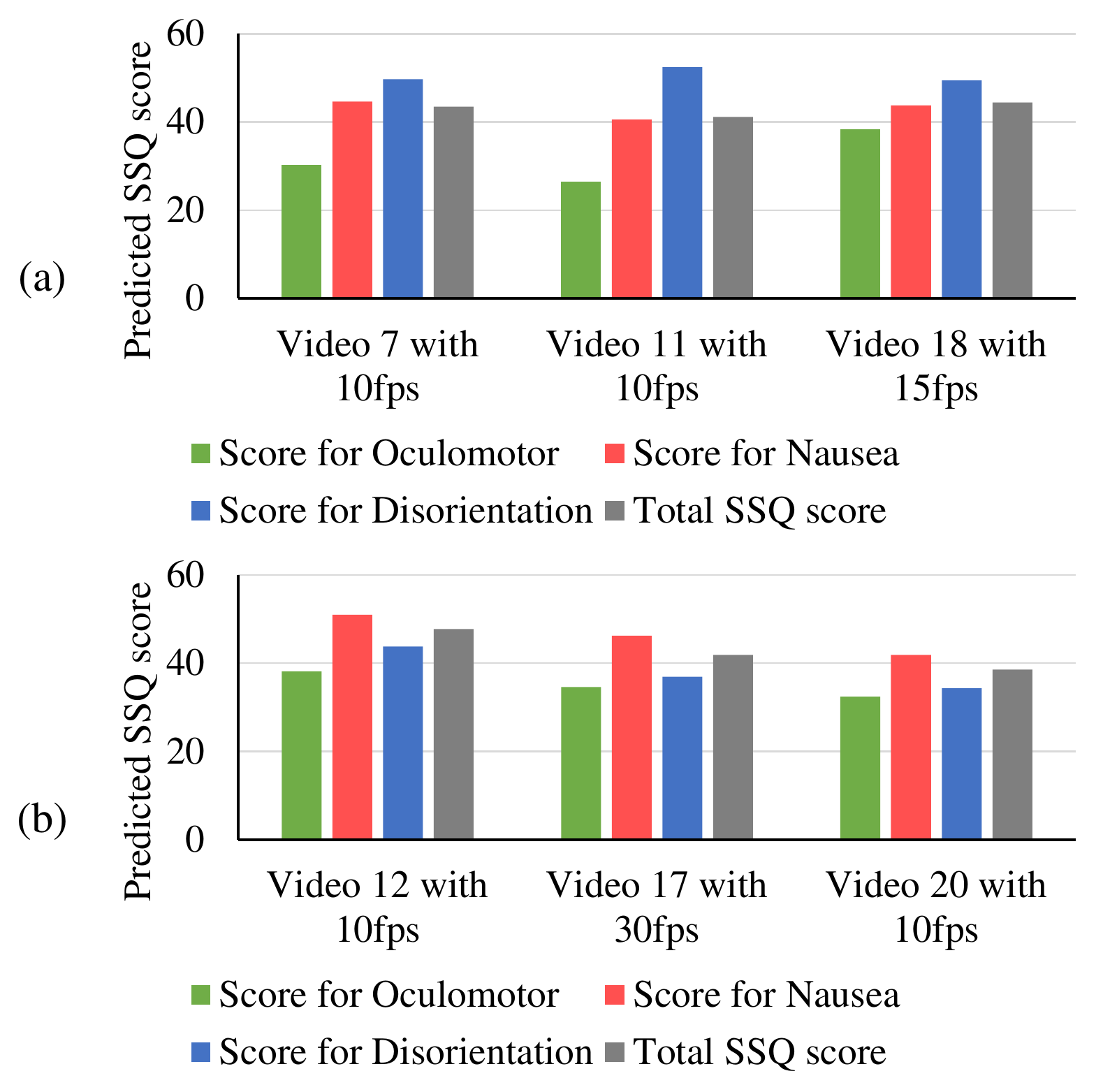}
\end{center}
   \caption{Predicted physical symptom scores for VR contents with different characteristics. (a) Examples of VR contents causing severe disorientation and (b) Examples of VR contents causing severe nausea. The green bar and red bar represent the symptom scores for oculomotor and nausea, respectively. The blue bar represents the disorientation score. The gray bar means a total VR sickness score.}
\label{fig:4}

\end{figure}

Figure~\ref{fig:4} shows the predicted physical symptom scores by our method for various datasets. In Figure~\ref{fig:4}(a), 360-degree videos (Video 7, 11, 18) include acceleration and rapid rotation in various directions and low frame rate. In case of that, the subjects scored high on disorientation symptoms. It means that they mainly felt dizziness, vertigo, and fullness of head categorized into the disorientation symptoms. In particular, the proposed method could well-predict each physical symptom score. As a result, the total SSQ score for an overall degree of VR sickness is well estimated in TABLE 3. In Figure~\ref{fig:4}(b), the 360-degree videos (Video 12, 17, 20) are not fast but contains a lot of shaking. In Figure~\ref{fig:4}(b), the physical symptom scores indicate that the subjects felt severe nausea symptom, compared to disorientation and oculomotor when they watched the contents with extreme shaking. The proposed method could reliably provide the degree of each physical symptom of VR sickness for a given VR content. It can be useful as a VR viewing safety guideline tool because the proposed model can provide useful information about what physical symptoms will cause.

%---------------------------------------------Fig.4
%------------------------------------ Figure 4

%--------------------------------------------------

In addition, we evaluated the performance of overall VRSA. In this experiment, we commonly compared with two physiological signal-based methods \cite{11,12} using the collected HR and GSR. The standard deviation of the HR in time domain was used as objective metric for heart rate variability (HRV)-based method. For the GSR-based method, the mean of GSR value in time domain was used as an objective metric. In addition, we performed VR sickness assessment by measuring the optical flow of 360-degree videos \cite{37}. The average magnitude and direction of optical flow were used as a VR sickness feature. On VRPS DB-FR, the frame rate value was used as an objective metric as well. These performance metrics were calculated by the non-linear regression using logistic function \cite{33}. In addition, the VR sickness score was assessed using deep learning-based method with 3D CNN, which has the same architecture of our discriminator. TABLE 2 shows the prediction performance for our method and other methods on the public VRSA dataset \cite{1}. On \cite{1}, the GSR-based method \cite{11} and the optical flow-based model achieved a good correlation. Our model except for comparison network (‘NS+PSP’) provided about 70\% correlation as well. In particular, our model (‘NS+C+PSP’) had the highest correlation with subjective sickness score (PLCC was 0.891 and SROCC was 0.882). The RMSE value of our ‘NS+C+PSP’ was significantly lower than those of the existing methods.

TABLE 3 shows the prediction performance evaluation on VRPS DB-FR. Our model (‘NS+C+PSP’) was highly correlated with subjective scores on the dataset with different characteristics (PLCC: 0.831, SROCC: 0.812, and RMSE: 7.112), compare to the objective VRSA methods using physiological measurement, frame rate value, conventional 3D CNN, and the state-of-the-art VRSA \cite{19}. Even one of the recent works, VRSA Net \cite{19}, could not work well on our VRPS DB-FR datasets because the characteristics of our dataset are different from those of KAIST IVY 360-degree video DB. These results indicate that the proposed physical symptom prediction model can be utilized to evaluate the level of VR sickness for VR content with various characteristics. In addition, the ablation study of our model in TABLE 1, 2 and 3 showed that the encoded neural mismatch feature is effective to estimate physical symptoms and overall VR sickness.

%-------------------------------------------Table 2
%------------------------------------------TABLE 2.
\begin{table}[t!]
  \centering
%  \begin{adjustbox}{width=0.55\textwidth}
  \label{Table:2}
  \resizebox{\columnwidth}{!}{% resize to the page column width
\begin{tabular}{c|c|c|c}
\hline
\hline
\textbf{Objective metrics} & \textbf{PLCC}  & \textbf{SROCC} & \textbf{RMSE}  \\ \hline
HRV-based method           & 0.515          & 0.400          & 21.172         \\ \hline
GSR-based method           & 0.691          & 0.695          & 20.405         \\ \hline
Optical flow-based method  & 0.717          & 0.710          & 15.316         \\ \hline
3D CNN-based method 	   & 0.612          & 0.568          & 20.901         \\ \hline
VRSA Net \cite{19}		   & 0.885          & 0.882          & 10.251         \\ \hline
NS+PSP	(ours)				   & 0.695          & 0.661          & 11.928         \\ \hline
\textbf{NS+C+PSP (ours)}     & \textbf{0.891} & \textbf{0.882} & \textbf{9.651} \\ 
\hline
\hline
\end{tabular}
}
%\end{adjustbox}
  \centering
  \caption{Prediction Performance on KAIST IVY 360-Degree Video DB \cite{1}}
\end{table}

%--------------------------------------------------
%-------------------------------------------Table 3
%------------------------------------------TABLE 3.
\begin{table}[t!]

  \centering
%    \begin{adjustbox}{width=0.55\textwidth}
  \label{Table:3}
    \resizebox{\columnwidth}{!}{% resize to the page column width
\begin{tabular}{c|c|c|c}
\hline
\hline
\textbf{Objective metrics} & \textbf{PLCC}  & \textbf{SROCC} & \textbf{RMSE}  \\ \hline
HRV-based method         & 0.505          & 0.514          & 10.078         \\ \hline
GSR-based method         & 0.395          & 0.351          & 12.721         \\ \hline
Frame rate-based method  & 0.454          & 0.538          & 11.901         \\ \hline
3D CNN-based method 	 & 0.595          & 0.548          & 10.136         \\ \hline
VRSA Net \cite{19}		 & 0.624          & 0.601          & 9.351         \\ \hline
NS+PSP	(ours)			 & 0.651          & 0.609          & 10.742         \\ \hline
\textbf{NS+C+PSP	(ours)}     & \textbf{0.831} & \textbf{0.812} & \textbf{7.112} \\ 
\hline
\hline
\end{tabular}
%  \end{adjustbox}
}
  \centering
\caption{Prediction Performance on 360-Degree Video Database with Different Frame Rates (our VRPS DB-FR)}
\end{table}

%--------------------------------------------------

%===================================================================================
%								6. CONCLUSION
%===================================================================================

\section{Conclusion}
In this paper, we proposed a novel objective physical symptom prediction to make better understanding of VR sickness. We addressed a limitation of existing works that did not consider the physical symptoms. Furthermore, we built eighty 360-degree videos with four different frame rates and conducted extensive subjective experiments to obtain physiological signals (HR and GSR) and subjective questionnaire (SSQ scores) for physical symptom scores. In extensive experiments, we demonstrated that our model could provide not only overall VR sickness score but also the levels of physical symptoms of VR sickness. This can be utilized as practical applications for viewing safety of VR contents.

%===================================================================================
%								ACKNOWLEDGEMENTS
%=================================================================================
\section{Acknowledgements}
This work was partly supported by IITP grant (No. 2017-0-00780), IITP grant (No. 2017-0-01779), and BK 21 Plus project. H.T. Lim is now in NCSOFT, Korea.

%===================================================================================
%								   REFERENCES
%===================================================================================

\bibliography{refs_aaai_vr}

\end{document}